\documentclass[runningheads]{llncs}
\usepackage[T1]{fontenc}
\usepackage{graphicx,verbatim}
\usepackage{amsmath}
\usepackage{amssymb}
\begin{document}
%
\title{Decomposing Whole Slide Image Report Generation with Graph-Constrained Multiple Instance Learning Workflows}

\titlerunning{Decomposing Whole Slide Image Report Generation}
%
\author{Antony Gitau\inst{1, 2} \and
Martyna Borak\inst{3} \and
Bjørn-Jostein Singstad\inst{1,4}\and
Martin Paulson\inst{2}\and
Karl Thomas Hjelmervik\inst{1}\and
Ola Marius Lysaker\inst{1}\and
Veralia Gabriela Sanchez\inst{1}}
\authorrunning{A. Gitau et al.}
%
\institute{Faculty of Technology, Natural Sciences and Maritime Sciences, University of South-Eastern Norway, Borre, Norway; \email{antony.gitau@usn.no} \and
Center for Cancer and Blood Diseases, Vestfold Hospital Trust, Tønsberg, Norway \and
Faculty of Biomedical Engineering, Silesian University of Technology, Zabrze, Poland \and
Department of Radiology, Vestfold Hospital Trust, Tønsberg, Norway 
}
  
\maketitle              
\begin{abstract}

Whole-slide image (WSI) report generation requires recognizing spatially distributed pathological features and organizing them into a coherent diagnostic narrative. Although direct vision-to-text models can yield fluent reports, they obscure the contributions and failure modes of visual recognition, structured reasoning, and language generation. We propose a decomposed framework in which frozen Virchow2 tile embeddings are aggregated by multiple-instance learning (MIL) classification heads that answer organ-specific diagnostic questions. An organ-conditioned graph constrains the assembly of these answers into a structured reasoning chain, which a language model realizes as a pathology report. On the REG2026 held-out set of 2,028 slides, the proposed workflow achieved a chain-Jaccard score of 0.702. Performance fell to 0.420 without graph-based chain construction, 0.398 when the organ-specific graphs were replaced by a single organ-agnostic graph, and 0.371 when the language model constructed the chain freely from MIL predictions. Using the same report generator, graph-structured chains improved the report score from 0.330 to 0.495. On 350 external TCGA WSIs spanning the seven REG organs without fine-tuning, the expected organ graph was selected in 64.0\% of cases and ranked among the top three in 86.6\%. Providing the correct organ graph increased agreement with coarse TCGA primary-diagnosis labels from 61.8\% to 92.6\%, identifying organ routing as a main bottleneck under domain shift. Overall, organ-conditioned, graph-constrained chain assembly improves structured reasoning and report generation while enabling stage-specific error localization.


\end{abstract}
\section{Introduction}
Generating pathology reports from whole-slide images (WSIs) requires models
to analyze gigapixel-scale tissue and produce clinically structured diagnostic
text. Recent WSI report-generation methods aggregate patch-level visual
representations and decode them into reports, sometimes using regional feature
selection or retrieval from similar cases \cite{chen2024wsicaption,hu2025pathology}. Related work in medical report
generation has explored structured or modular representations to improve the
organization and consistency of generated reports
\cite{li2019knowledge,delbrouck2025automated,kim2025enhancing}. However,
these approaches still largely couple visual recognition, diagnostic reasoning,
and language realization. Consequently, an incorrect report does not readily
reveal whether the failure arose from missed visual evidence, an inappropriate
diagnostic pathway, or inaccurate textual realization limiting interpretability.

We therefore decompose WSI report generation into three explicit stages: visual evidence prediction, graph-constrained diagnostic reasoning, and report realization. First, a frozen Virchow2 \cite{zimmermann2024virchow2} encoder extracts tile-level representations, which are aggregated by MIL \cite{ilse2018attention} heads to answer bounded diagnostic questions. The predicted organ selects a corresponding diagnostic graph, whose canonical questions, answer vocabularies, and allowable transitions determine how these predictions are assembled into a reasoning chain. Finally, Qwen2.5-1.5B-Instruct \cite{qwen2024qwen25} converts the structured chain into a pathology report. In this design, MIL provides slide-level visual evidence, the graph enforces organ-specific diagnostic structure, and the language model is used primarily for report realization.

We evaluate the method on REG2026 \cite{song2026reg2026} using a held-out split and show that the full graph-constrained workflow outperforms ablations that remove visual evidence, graph structure, or organ conditioning. We also show that reports generated from graph-assembled reasoning chains outperform direct generation from unstructured MIL. Finally, on 350 external TCGA diagnostic WSIs \cite{grossman2016gdc}, we find that organ routing
is the principal bottleneck under domain shift.

Contributions can be summarized as follows:

\begin{itemize}
    \item We propose a graph-constrained MIL workflow for WSI report generation that converts tile-level visual evidence into organ-specific reasoning chains before language-model report realization.

    \item We present a decomposition analysis on REG2026 showing that visual evidence, graph structure, and organ conditioning each contribute to reasoning performance, and that graph-structured chains improve report generation compared with unstructured visual facts.

    \item We perform an external analysis on 350 TCGA WSIs  using organ routing and weak diagnostic concordance to localize domain-shift failures to workflow selection rather than within-organ diagnostic reasoning.
\end{itemize}

\section{Methods and materials}
\paragraph{\textbf{Dataset.}}
We used the REG2026 training set as the primary development and evaluation resource. The dataset contains 11,220 WSIs with organ labels, chain-of-thought (CoT) annotations, and diagnostic reports across seven organ-specific workflows: urinary bladder, breast, uterine cervix, colon, lung, prostate, and stomach. 

For internal evaluation, we constructed a train split of 8,979 slides and 2,028 held out test after dropping 213 slides from the held out slides as they shared a case identifier with a slide in the train split. 

For external analysis, we sampled 350 public diagnostic TCGA WSIs, with 50 WSIs per REG-compatible organ project: TCGA-BLCA, TCGA-BRCA, TCGA-CESC, TCGA-COAD, TCGA-LUAD, TCGA-PRAD, and TCGA-STAD. Since TCGA does not provide REG-style CoT annotations or reference reports, we evaluated workflow routing and weak diagnostic concordance against Genomic Data Commons (GDC) primary-diagnosis metadata.

\begin{figure}[hbp!]
\includegraphics[width=\textwidth]{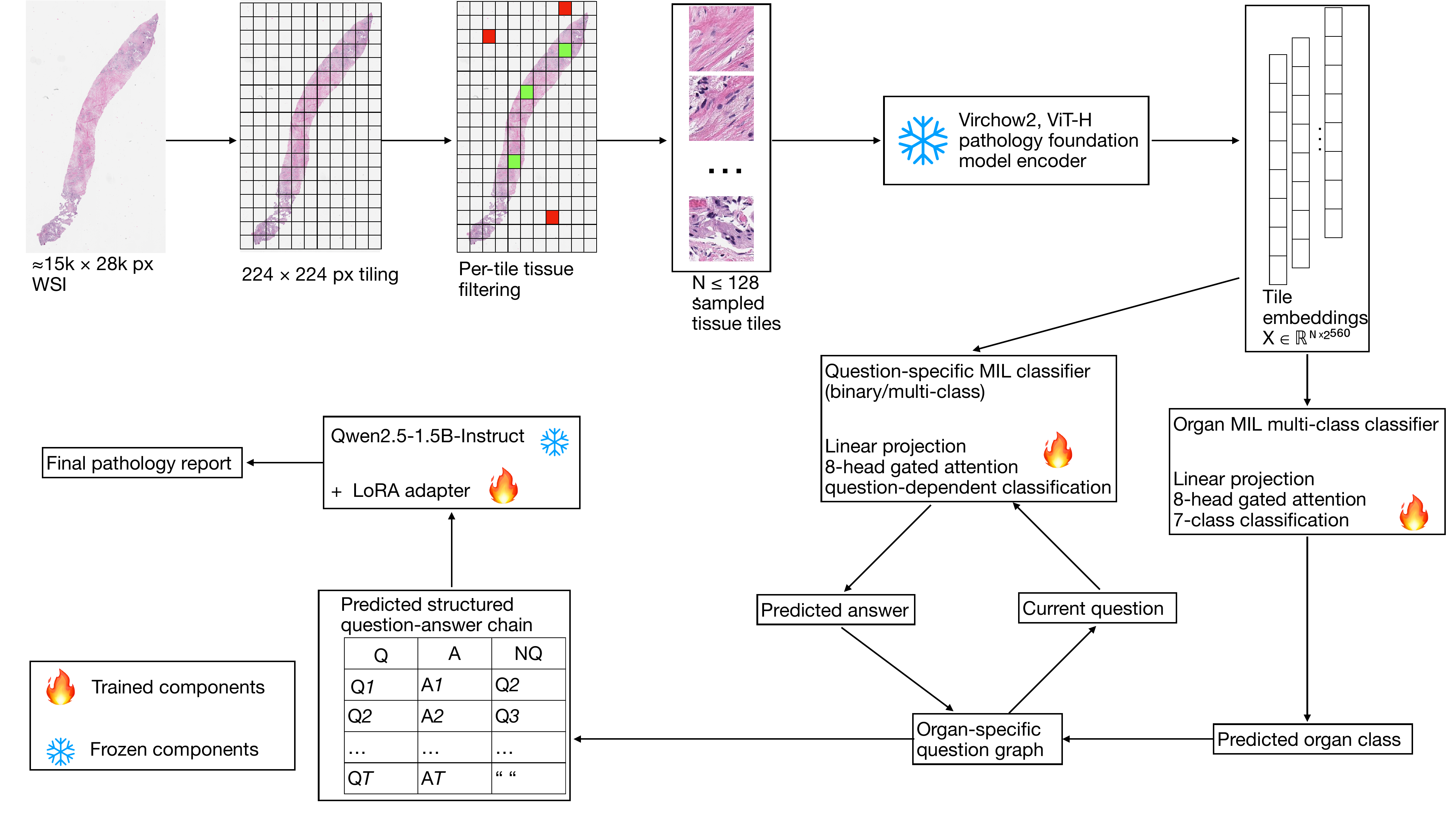}
\caption{Overview of the proposed framework. The WSI is densely divided into non-overlapping $224  \times 224$ pixel candidate tiles. Tiles are retained when at least 50\% of their pixels have a mean RGB intensity below 220, and up to 128 retained tissue tiles are uniformly and randomly sampled during training and inference. A frozen Virchow2 encoder transforms these tiles into an embedding bag used by an organ MIL classifier, which selects the organ-specific reasoning graph, and organ–question-specific MIL classifiers, which predict answers to successive graph questions. Graph-guided reasoning organizes these predictions into a structured question–answer chain, which Qwen2.5-1.5B-Instruct with LoRA converts into the final pathology report. } \label{fig:workflow}
\end{figure}

\paragraph{\textbf{Workflow overview.}} The proposed workflow is summarized in Figure~\ref{fig:workflow}. The accompanying code is publicly available on GitHub\footnote{
\url{https://github.com/Antony-gitau/SiV-USN-REG2026}}.

\paragraph{\textbf{Tile encoding and visual prediction.}}
For each slide, we extract tissue-containing $224 \times 224$ RGB tiles at the $20\times$, capping each bag at 128 embeddings. Training and evaluation sample identically with a fixed seed; the submitted container sampled more sparsely to meet its inference budget. Each tile is encoded with a frozen Virchow2 pathology foundation model. Let $O_i \in \mathbb{R}^{L \times d}$ denote the token output of Virchow2 for tile $i$, where $L$ is the number of output tokens and $d$ is the token dimension. We form the tile representation by concatenating the class-token embedding with the mean embedding over the remaining patch tokens after the initial special/register tokens:
\begin{equation}
x_i =
\left[
O_{i,0};
\frac{1}{L-5}\sum_{\ell=5}^{L-1} O_{i,\ell}
\right]
\in \mathbb{R}^{2560}.
\label{eq:tile_embedding}
\end{equation}
$O_{i,0}$ denotes the class token, and the second term corresponds to the averaged patch-token representation used by the implementation. A slide $s$ is then represented as a variable-length bag of tile embeddings,
\begin{equation}
X_s = \{x_{s,i}\}_{i=1}^{N_s}, \quad N_s \leq 128,
\label{eq:slide_bag}
\end{equation}
rather than as a single mean-pooled slide vector.

Each visual decision module applies gated attention MIL \cite{ilse2018attention} to obtain an attention-weighted slide representation from $X_s$. A classification head then predicts either the organ label or the answer to an organ-specific diagnostic question. Separate question heads are trained
only on slides whose reference chains contain the corresponding question.

\paragraph{\textbf{Organ-conditioned chain assembly.}}
The structured reasoning space is defined by organ-specific graphs from the REG2026 CoT annotations. For each organ $o$, we construct a graph
\begin{equation}
G_o = (V_o, E_o, A_o, T_o),
\label{eq:organ_graph}
\end{equation}
where $V_o$ is the set of canonical diagnostic question nodes, $E_o$ is the set of observed answer-conditioned transitions, $A_o$ stores the answer vocabularies, and $T_o$ is the transition function. 

At inference, an organ router first selects the workflow:
\begin{equation}
\hat{o}_s = \arg\max_o p_{\theta_{\mathrm{org}}}(o \mid X_s).
\label{eq:organ_router}
\end{equation}
The selected graph $G_{\hat{o}_s}$ is traversed from its start node. At step $t$, the system visits question $q_t$ and predicts an answer using the corresponding question-specific MIL classifier head:
\begin{equation}
\hat{a}_t =
\arg\max_{a \in A_{\hat{o}_s,q_t}}
p_{\theta_{\hat{o}_s,q_t}}(a \mid X_s).
\label{eq:answer_prediction}
\end{equation}
The graph determines the next question as
$q_{t+1}=T_{\hat{o}_s}(q_t,\hat{a}_t)$. Repeating this procedure produces the
ordered chain:
\begin{equation}
\hat{C}_s =
\left((q_t,\hat{a}_t,q_{t+1})\right)_{t=1}^{T_s}.
\label{eq:predicted_chain}
\end{equation}

The organ router and all question-specific MIL heads are optimized with cross-entropy, while the graphs are deterministic structures extracted from the training workflows. The final submitted model uses all 11,220 REG2026 training WSIs.

\paragraph{\textbf{Report realization.}}
Report generation is conditioned on the predicted reasoning chain rather than directly on image features. We fine-tuned Qwen2.5-1.5B-Instruct with LoRA adapters on the reference structured findings using the standard autoregressive language-model objective.

\subsection{Experiments}
We tested whether visual evidence, graph-constrained reasoning, organ-conditioned workflow selection, and language-model report realization each contributed distinct information to WSI report generation. 
\paragraph{\textbf{Ablations.}}
The \emph{Full} model uses MIL classifiers to answer graph questions and assembles the reasoning chain using the selected organ-specific graph. The \emph{Majority} traverses the ground-truth organ's graph and replaces each visual prediction with the training-set majority answer, isolating the value of visual evidence given correct routing. The \emph{Heads, no graph} emits the answers predicted by the MIL heads without graph traversal, preserving visual prediction while removing structured reasoning. The \emph{No organ routing} uses the visual heads but traverses a single organ-agnostic graph formed from the union of all seven organ graphs, removing organ-conditioned workflow selection. The \emph{Free-form} gives Qwen2.5-1.5B-Instruct the same MIL answers and asks it to assemble the reasoning chain without exposing the graph schema. The \textit{Canonical template} follows the transition associated with each node's training-set modal answer, making question selection independent of the slide-specific predicted answer. \textit{Shuffled transitions} permutes the answer-to-transition mapping within each branching node while preserving its outgoing-edge set. The last two use the \emph{Full} model's predicted organ, vocabularies, graph nodes, and fixed MIL heads.

We also evaluated the same report realizer under three input conditions. The \emph{GT-chain} received the reference CoT, providing an approximate upper bound for report realization when reasoning is correct. The \emph{Predicted-chain} received the chain produced by our graph-constrained MIL workflow. The \emph{Facts} received the flat set of MIL predictions without graph-constrained assembly. Because the report generator is unchanged, differences primarily reflect the quality and structure of its conditioning information.

\paragraph{\textbf{Evaluation metrics.}}
For each predicted reasoning chain $\hat{C}_s$ and reference chain $C_s$, we
computed chain-level Jaccard similarity over question-answer pairs:
\begin{equation}
J(C_s,\hat{C}_s)
=
\frac{
|\mathcal{P}(C_s) \cap \mathcal{P}(\hat{C}_s)|
}{
|\mathcal{P}(C_s) \cup \mathcal{P}(\hat{C}_s)|
},
\label{eq:chain_jaccard}
\end{equation}
where $\mathcal{P}(C)$ denotes the set of
$(\mathrm{question},\mathrm{answer})$ pairs in chain $C$.  We additionally measure organ-routing accuracy, per-question answer accuracy, vocabulary validity, and answer-transition consistency.

For challenge-aligned evaluation, we report the official REG2026 Workflow
Reasoning score (Metric A) \cite{song2026reg2026}. It combines exact path
agreement (BPV), partial overlap between directed question transitions
(Edge-F1), semantic similarity of intermediate answers over reference
non-final edges (MESS), and the final-report submetric. The report submetric combines biomedical-keyword Jaccard, ROUGE-L, BLEU-4, and PubMedBERT embedding similarity using the
organizer-defined preprocessing and normalization. The per-case score is defined by Eq. \ref{eq:workflow_a} and the dataset score is the mean across cases.
\begin{equation}
\mathrm{WorkflowA}
=
0.05\,\mathrm{BPV}
+
0.30\,\mathrm{EdgeF1}
+
0.25\,\mathrm{MESS}
+
0.40\,\mathrm{Report},
\label{eq:workflow_a}
\end{equation}

\section{Results}
The full graph-constrained workflow was the best configuration in every organ (Table~\ref{tab:per_organ_reasoning}), reaching a chain-Jaccard of 0.702 with organ routing accuracy of 0.987 (0.968–1.000 per organ) and per-question answer accuracy 0.813. It exceeded the Canonical template (0.572) by 0.129 (paired-bootstrap 95\% CI [0.120, 0.139]) and the Shuffled-transitions control (0.320) by 0.382 ([0.373, 0.391]). 

\begin{table}
\caption{Per-organ reasoning performance on the held-out set; all values
report chain-Jaccard. Canon uses modal-answer transitions, whereas Shuf
permutes answer-to-transition mappings within branching nodes.}
\label{tab:per_organ_reasoning}
\centering
{\fontsize{8}{9}\selectfont
\setlength{\tabcolsep}{2.4pt}
\renewcommand{\arraystretch}{1.05}
\begin{tabular}{|l|r|c|c|c|c|c|c|c|}
\hline
\textbf{Organ} & \textbf{$n$} & \textbf{Full} & \textbf{Canon} &
\textbf{Shuf} & \textbf{Majority} & \textbf{Heads} &
\textbf{No organ} & \textbf{Free-form} \\
\hline
Bladder  & 153 & \textbf{0.678} & 0.616 & 0.318 & 0.512 & 0.547 & 0.457 & 0.490 \\
Breast   & 442 & \textbf{0.633} & 0.510 & 0.236 & 0.484 & 0.349 & 0.403 & 0.313 \\
Cervix   & 162 & \textbf{0.737} & 0.733 & 0.390 & 0.640 & 0.562 & 0.359 & 0.448 \\
Colon    & 399 & \textbf{0.690} & 0.552 & 0.337 & 0.426 & 0.425 & 0.378 & 0.370 \\
Lung     & 188 & \textbf{0.773} & 0.734 & 0.374 & 0.632 & 0.510 & 0.391 & 0.410 \\
Prostate & 333 & \textbf{0.738} & 0.479 & 0.363 & 0.401 & 0.391 & 0.476 & 0.391 \\
Stomach  & 351 & \textbf{0.723} & 0.581 & 0.304 & 0.472 & 0.358 & 0.338 & 0.317 \\
\hline
\textbf{All} & \textbf{2028} & \textbf{0.702} & 0.572 &
0.320 & 0.485 & 0.420 & 0.398 & 0.371 \\
\hline
\end{tabular}}
\end{table}

Using the same Qwen2.5-1.5B-Instruct report realizer across conditions: conditioning on our graph-assembled chain increased the report submetric from 0.330 with flat MIL-predicted findings to 0.495. The improvement was positive for all organs ($+0.020$ to $+0.284$). Conditioning on the reference chain instead of our predicted chain raised the report score from 0.495 to 0.658, attributing 0.163 to chain errors and the remaining gap to realization and metric strictness.

To localize errors within the graph-constrained workflow, we grouped the 175 organ-specific question nodes encountered in the held-out test into five task categories informed by common elements of synoptic pathology reporting \cite{srigley2021iccr}. The residual concentrated in fine-grained attributes (Table~\ref{tab:question_category_errors}). Grading/quantification questions were least accurate (0.686) and, though only 22 nodes, produced the largest error share (38.0\%), while diagnosis/subtype questions (0.849) contributed a comparable 36.9\% through frequency; presence/extent (0.942) and workflow/schema (0.963) were reliable.

\begin{table}
\caption{Post-hoc category-level analysis of question-node errors on the
held-out set. Error share denotes the proportion of all incorrect
question-level predictions attributable to each category.}
\label{tab:question_category_errors}
\centering
{\fontsize{8}{9}\selectfont
\setlength{\tabcolsep}{4.0pt}
\renewcommand{\arraystretch}{1.08}
\begin{tabular}{|l|r|r|c|c|}
\hline
\textbf{Question category} &
\textbf{Nodes} &
\textbf{Occurrences} &
\textbf{Accuracy} &
\textbf{Error share} \\
\hline
Grading / quantification & 22 & 3,689 & 0.686 & 38.0\% \\
Diagnosis / subtype      & 49 & 7,437 & 0.849 & 36.9\% \\
Presence / extent        & 68 & 8,274 & 0.942 & 15.9\% \\
Workflow / schema        & 31 & 7,252 & 0.963 & 8.8\% \\
Rare findings           & 5  & 83    & 0.867 & 0.4\% \\
\hline
\textbf{Total} &
\textbf{175} &
\textbf{26,735} &
-- &
\textbf{100\%} \\
\hline
\end{tabular}}
\end{table}

On the held-out set, the full model scored Workflow-A 0.725 (BPV 0.592, Edge-F1 0.914, MESS 0.894, report 0.495), exceeding the Canonical (Edge-F1 0.761) and Shuffled (0.469) controls. Full also exceeded Canonical on both edge precision (0.919 vs.\ 0.730) and recall (0.923 vs.\ 0.842), so its gain was not merely emitting fewer edges. On the Test-Phase-1 (TP1) set, the submitted container scored 0.683 overall, comprising Workflow-A 0.558 (BPV 0.223, Edge-F1 0.768, MESS 0.703, report 0.352) and Visual-Grounding-B 0.975. Although the held-out and TP1 evaluations use different models, datasets, and tile sampling, no single graded component collapses. Edge-F1 retains 84\% of its held-out value, MESS 79\%, and the report 71\%. BPV retains 38\%, since as an exact edge-set match, it amplifies small errors. For example, over a chain of ~16 edges, per-edge accuracies of 0.97 and 0.91 give BPV values of 0.61 and 0.22. Finally, on 350 external TCGA slides, top-1 organ routing decreased to 0.640 under domain shift. Providing the correct organ graph increased agreement with coarse GDC primary-diagnosis labels from 0.618 to 0.926 (Table~\ref{tab:tcga_external}).

\begin{table}
\caption{External evaluation on 350 TCGA diagnostic WSIs, with 50 slides
per organ. Route denotes top-1 organ-routing accuracy and Top-3 indicates
whether the expected organ was among the three highest-ranked predictions.
Weak concordance measures agreement with coarse GDC primary-diagnosis labels;
Oracle reports concordance when the expected organ graph is provided.}
\label{tab:tcga_external}
\centering
{\fontsize{8}{9}\selectfont
\setlength{\tabcolsep}{2.0pt}
\renewcommand{\arraystretch}{1.08}
\begin{tabular}{|l|r|c|c|c|c|}
\hline
\textbf{Organ} &
\textbf{$n$} &
\textbf{Route} &
\shortstack{\textbf{Expected organ}\\\textbf{in top 3}} &
\shortstack{\textbf{Weak}\\\textbf{concordance}} &
\shortstack{\textbf{Oracle-organ}\\\textbf{concordance}} \\
\hline
Bladder  & 50 & 11/50 (0.220) & 29/50 (0.580) & 8/42 (0.190)  & 30/42 (0.714) \\
Breast   & 50 & 42/50 (0.840) & 49/50 (0.980) & 37/49 (0.755) & 44/49 (0.898) \\
Cervix   & 50 & 17/50 (0.340) & 33/50 (0.660) & 15/49 (0.306) & 47/49 (0.959) \\
Colon    & 50 & 38/50 (0.760) & 50/50 (1.000) & 35/50 (0.700) & 45/50 (0.900) \\
Lung     & 50 & 47/50 (0.940) & 50/50 (1.000) & 47/50 (0.940) & 50/50 (1.000) \\
Prostate & 50 & 46/50 (0.920) & 48/50 (0.960) & 45/50 (0.900) & 49/50 (0.980) \\
Stomach  & 50 & 23/50 (0.460) & 44/50 (0.880) & 23/50 (0.460) & 50/50 (1.000) \\
\hline
\textbf{All} &
\textbf{350} &
\textbf{224/350 (0.640)} &
\textbf{303/350 (0.866)} &
\textbf{210/340 (0.618)} &
\textbf{315/340 (0.926)} \\
\hline
\end{tabular}}
\end{table}

\section{Discussion}
Decomposing WSI report generation into visual prediction, organ-conditioned workflow assembly, and language realization improved performance while making failure sources observable. The full workflow outperformed every ablation, showing that visual evidence, organ conditioning and diagnostic structure provide complementary information. Canonical and Shuffled used the same predicted organ, vocabularies, graph nodes, and MIL answers as the full model, but achieved lower chain-Jaccard, indicating graph contribution beyond mere formatting. Conditioning the same realizer on the graph-assembled chain rather than flat MIL findings increased the report score, indicating that selection and ordering of findings
provide a more effective intermediate representation than an unordered fact set.

Error analysis localizes that residual to fine-grained attributes: grading and quantification, and diagnosis and subtype questions were least accurate and together produced ~75\% of node-level errors, whereas presence and workflow questions were reliable. This pattern is clinically plausible since grading requires subtle, focal, or integrative morphologic assessment. The external TCGA stress test further shows the decomposition's diagnostic value: end-to-end concordance fell under domain shift, but supplying the correct organ graph recovered it (0.62→0.93), identifying external organ routing, not within-organ prediction, as an important contributor to error under this coarse weak-label evaluation. 

\paragraph{\textbf{Limitations and future work.}} The graphs are derived from REG2026 annotations and may not represent unseen workflows. Additionally, the frozen Virchow2 encoder with MIL heads and the 128-tile cap may miss subtle grading cues or sparse diagnostic regions. Future work should target the two identified failure points: external organ routing through calibrated top-$k$ routing or domain adaptation, and fine-grained attribute prediction through multi-scale or uncertainty-guided tiling and hard-example training. Future evaluation should also assess factual correctness on paired WSI–report datasets.

\section{Conclusion}

We presented a graph-constrained MIL workflow for reasoning-guided WSI report generation. By decomposing the task into frozen tile encoding, organ-conditioned graph traversal with question-specific MIL heads, and language-model report realization, the method produces reasoning chains and interpretable intermediate predictions. On a REG2026 held-out set, the full workflow outperformed ablations that removed visual evidence, graph structure, or organ conditioning. Structured reasoning also improved final report generation compared with flat visual facts. External TCGA evaluation further showed that the decomposition can localize domain-shift failures, separating organ-routing errors from within-organ diagnostic reasoning. Overall, the results support graph-constrained visual reasoning as a practical and interpretable route toward WSI pathology report generation. 

\section{Disclosure of Interests}
The authors have no competing interests to declare that are relevant to the content of this article.

\bibliographystyle{splncs04}
\bibliography{reference}
%




\end{document}